\documentclass[final,1p,times,nopreprintline]{elsarticle}

\usepackage{amssymb}
\usepackage{amsmath}
\usepackage{siunitx}
\usepackage{pdflscape}

\usepackage{todonotes}
\usepackage{hyperref} 
\usepackage{rotating}

\usepackage{makecell}
\usepackage{array}

\usepackage{float}

\journal{}

\begin{document}

\begin{frontmatter}

\title{Benchmarking Convolutional Neural Network and Graph Neural Network based Surrogate Models on a Real-World Car External Aerodynamics Dataset}

\author[label1]{Sam Jacob Jacob} 
\author[label1]{Markus Mrosek} 
\author[label1]{Carsten Othmer} %
\author[label2]{Harald Köstler} %

\affiliation[label1]{organization={Volkswagen AG},%
            city={Wolfsburg},
            country={Germany}}

\affiliation[label2]{organization={Friedrich-Alexander-Universität Erlangen-Nürnberg},%
            city={Erlangen},
            country={Germany}}

\begin{abstract}
Aerodynamic optimization is crucial for developing eco-friendly, aerodynamic, and stylish cars, which requires close collaboration between aerodynamicists and stylists, a collaboration impaired by the time-consuming nature of aerodynamic simulations. Surrogate models offer a viable solution to reduce this overhead, but they are untested in real-world aerodynamic datasets. We present a comparative evaluation of two surrogate modeling approaches for predicting drag on a real-world dataset: a Convolutional Neural Network (CNN) model that uses a signed distance field as input and a commercial tool based on Graph Neural Networks (GNN) that directly processes a surface mesh. In contrast to previous studies based on datasets created from parameterized geometries, our dataset comprises 343 geometries derived from 32 baseline vehicle geometries across five distinct car projects, reflecting the diverse, free-form modifications encountered in the typical vehicle development process. Our results show that the CNN-based method achieves a mean absolute error of 2.3 drag counts, while the GNN-based method achieves 3.8. Both methods achieve approximately 77\% accuracy in predicting the direction of drag change relative to the baseline geometry. While both methods effectively capture the broader trends between baseline groups (set of samples derived from a single baseline geometry), they struggle to varying extents in capturing the finer intra-baseline group variations. In summary, our findings suggest that aerodynamicists can effectively use both methods to predict drag in under two minutes, which is at least 600 times faster than performing a simulation. However, there remains room for improvement in capturing the finer details of the geometry.

\end{abstract}


\end{frontmatter}

\section{Introduction}
\label{sec1}

Aerodynamics plays a crucial role in the transition towards more environmentally friendly fleets, as its impact on the range of electric vehicles is becoming increasingly evident. However, designing an aerodynamic yet aesthetically pleasing vehicle requires close collaboration between aerodynamicists and stylists. This collaboration is currently impaired by the time-consuming and computationally expensive nature of high-fidelity aerodynamic simulations. In the industry, a typical high-fidelity simulation takes about 20 hours when using 960 CPU cores. The expensive computational cost leads to a slower and limited number of iterations between the aerodynamicists and stylists during the development of a car project. Consequently, aerodynamicists are forced to evaluate only a few variations based on trial and error or expert intuition, often leading to suboptimal drag coefficient ($c_d$) improvements. While automated optimization algorithms exist, they are computationally prohibitively expensive and cannot consider the aesthetics of the geometry.  

\begin{figure}[t]
  \centering
  \makebox[\textwidth]{\includegraphics[width=1.55\textwidth]{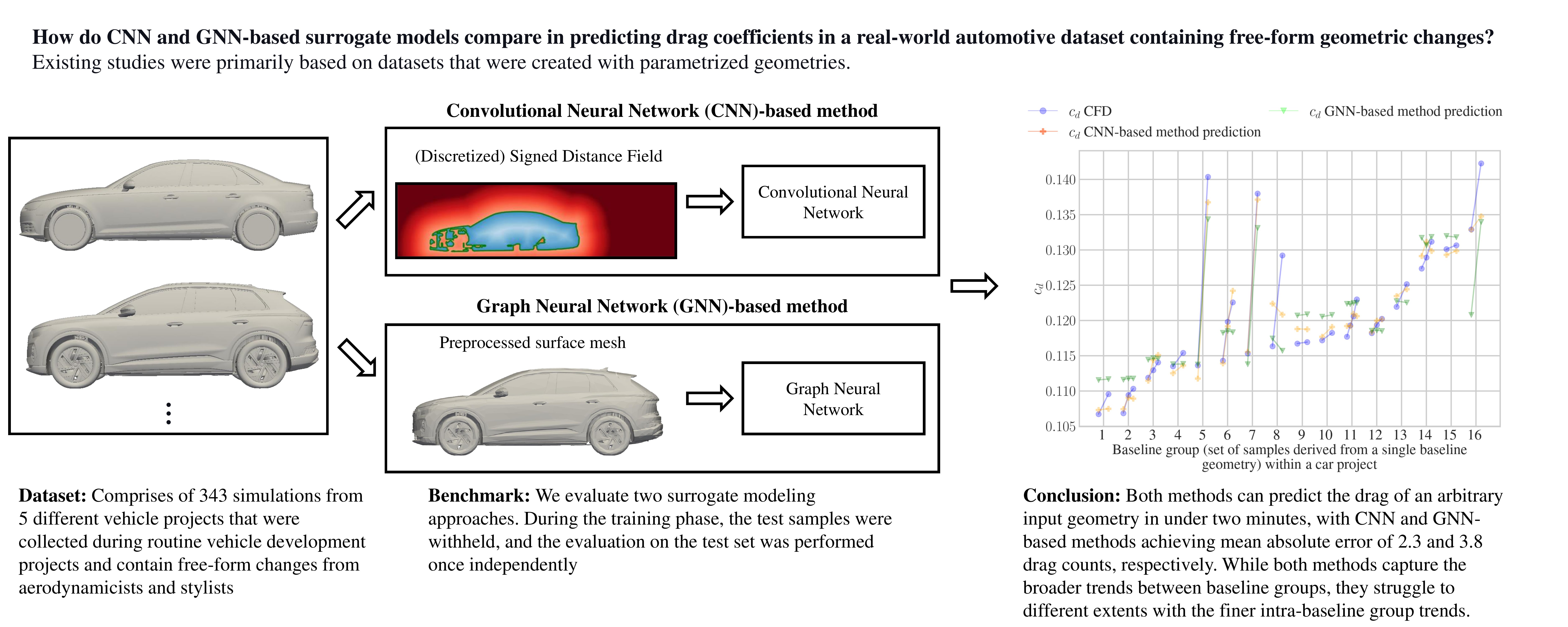}} 
  \caption{Graphical Abstract.}
  \label{fig:graphical_abstract}
\end{figure}

\subsection{Data-driven surrogate models}

In recent years, there has been a surge in studies that use machine learning methods to speed up the simulations or to approximate them, and an overview of these methods is beyond the scope of the paper and can be found in the review papers (e.g., \cite{Survery.Brunton.2020, Survery.Lino.2023}). Data-driven surrogate models \cite{Baque.2018, Jacob.2022, GNN.Elrefaie.3122024, GNN.Nabian.11262024, neuraloperator.Domino}, trained on simulations, has shown promise in quickly predicting the approximate $c_d$ of car geometries without any parameterization. This enables the aerodynamicists to explore a larger design space. 

We can use Kriging-based methods \cite{Kriging.Mrosek.2019} to predict the $c_d$, a parametric approach that requires the geometry to be parameterized and a corresponding design of experiment (DOE) for training. However, the adoption of Kriging-based methods and similar parametric approaches is limited as they restrict the predictions to the predefined parameter space and often require a new DOE when new geometric parameters are added to the geometry. In contrast, non-parametric approaches discussed below process the geometry directly without predefined parametrization and enable free-form changes.

Convolutional Neural Networks (CNNs) have been widely used for processing spatial data and feature extraction. However, a significant challenge in using CNNs for aerodynamic $c_d$ prediction is that the geometries are typically represented as meshes, which are inherently unstructured, and CNNs are natively designed for structured data. To overcome this limitation, two strategies are commonly employed to represent meshes in a structured grid:

\begin{itemize}
\item Binary Masks: Thuerey et al. \cite{Thuerey.2020} demonstrated that a binary mask (1 for the cells containing geometry and zero otherwise) generated on a Cartesian grid passed through a U-Net can predict airfoils' velocity fields. The disadvantage is that there is no global information about the geometry in the binary mask. 
\item Signed Distance Fields (SDF): SDF representation (see Section \ref{section:preprocessing} for more details) has both global and local information, and existing studies \cite{Jacob.2022, Guo.2016, SDF.Bhatnagar.2019, SDF.Chen., SDF.ThanhLuanTrinh.} have shown that models trained with SDF can handle arbitrary changes to the geometry. Guo et al. \cite{Guo.2016} showed that the global information from methods like SDF can improve the results compared to binary masks. One of the challenges is the generation of the SDFs, as complex or non-watertight geometries can produce artifacts or noise in the field; however, tools like OpenVDB \cite{OpenVDB} can generate robust SDFs even in these circumstances. 
\end{itemize}

Compared to other non-parametric approaches, a CNN-based method using SDF representation is relatively computationally cheaper during training and can be easily extended to predict volumetric fields.

Recent advances have seen the emergence of methods like Graph Neural Networks (GNN) and neural operators that can directly work on down-sampled meshes (as it is computationally expensive to train models on large meshes). Pfaff et al. \cite{GNN_base-pfaff2021learning} and Sanchez-Gonzalez et al. \cite{GNN_base-sanchez-gonzalez20a} successfully applied GNN for various simulation tasks, including flow around a cylinder and flow around a 2D airfoil. Further work on GNN \cite{GNN.Elrefaie.3122024, GNN.Nabian.11262024} and neural operators \cite{neuraloperator.Domino, neuraloperator.Li.Zongyi} showed that these methods could predict the aerodynamic $c_d$ of vehicles with good accuracy. One disadvantage of these methods is that they are computationally more expensive than CNN-based methods. However, they can directly work on meshes and enable the direct prediction of the fields on the surface of the geometry.

\subsection{Limitation of existing studies}

A critical limitation of existing studies on surrogate models for aerodynamic $c_d$ prediction is that they are evaluated on datasets that are well-suited for research but have significant shortcomings as they are not comparable to real-world usage; this limitation is due to difficulty in curating a dataset that simulates real-world usage. Most datasets are generated by parameterizing a limited number of geometries \cite{Jacob.2022, GNN.Elrefaie.3122024, Data.Elrefaie.6142024, drivaerAshton2024}, or using low-fidelity simulations \cite{GNN.Elrefaie.3122024, neuraloperator.Li.Zongyi}, or using geometries simpler than those used in the industry, often derived from geometries such as DrivAer \cite{DrivAer.Heft.2012} or ShapeNet \cite{shapenet2015}. These limitations mean the surrogate models are largely untested in real-world scenarios involving substantial free-form (non-parametric) geometric changes.

This study aims to bridge the gap by evaluating surrogate models on a dataset that closely imitates a real-world use case. None of the geometries or simulations were created explicitly for this dataset or research purposes; rather, the dataset comprises simulations collected during routine vehicle development projects by aerodynamicists. Our dataset features:
\begin{itemize}
\item Realistic geometries reflecting typical production vehicles from five different ongoing car projects.
\item High-fidelity simulations were performed using industry-standard solvers.
\item The dataset has 32 baseline vehicle geometries with free-form changes introduced by stylists and aerodynamics, capturing the inherent diversity in typical vehicle development projects.
\end{itemize}

\subsection{Benchmark}

In their current state, we believe that data-driven surrogate models only serve as a precursor to performing simulations and not as a replacement for simulations. We believe that aerodynamicists would typically explore a larger design space using surrogate models and then validate the most promising variations using traditional CFD simulation. The accurate prediction of $c_d$ is critical to finding the most promising variations during surrogate model exploration. So, in this study, we focus exclusively on predicting the $c_d$. We have performed a benchmark of two surrogate modeling approaches:

\begin{itemize}
\item A CNN-based model developed by the authors that uses an SDF as input (hereafter referred to as the CNN-based method).
\item A commercial tool\footnote{The company that developed this tool has requested to remain anonymous for this study.} based on GNN that uses a mesh as input (hereafter referred to as the GNN-based method).
\end{itemize}

The benchmarks were performed rigorously: 1) the models were trained by experts in the respective methods; 2) the test samples were withheld during training; 3) the evaluation of the test set was performed once independently. 

\subsection{Paper outline}

The study evaluates how well two surrogate modeling approaches perform in a real-world setting. The paper is organized as follows: We start with a description of the dataset in Section~\ref{sec:Dataset}, followed by the preprocessing methods, an online augmentation method for the CNN-based method, a comparison of the methods, and an overview of the evaluation metrics in Section~\ref{sec:Methodology}. Next, we have the results in Section~\ref{sec:Results}, where we compare not only the prediction accuracy of the methods but also other essential factors for the aerodynamicists, like the accuracy of the model in predicting the right direction of $c_d$ change relative to a baseline geometry and the ability to capture trends for a set of modifications performed on a baseline geometry. Finally, we conclude with a summary of our findings and suggest some directions for the future based on the observed limitations.

\section{Dataset} 
\label{sec:Dataset}

\begin{figure}[!htbp]
    \centering
    \includegraphics[width=\linewidth]{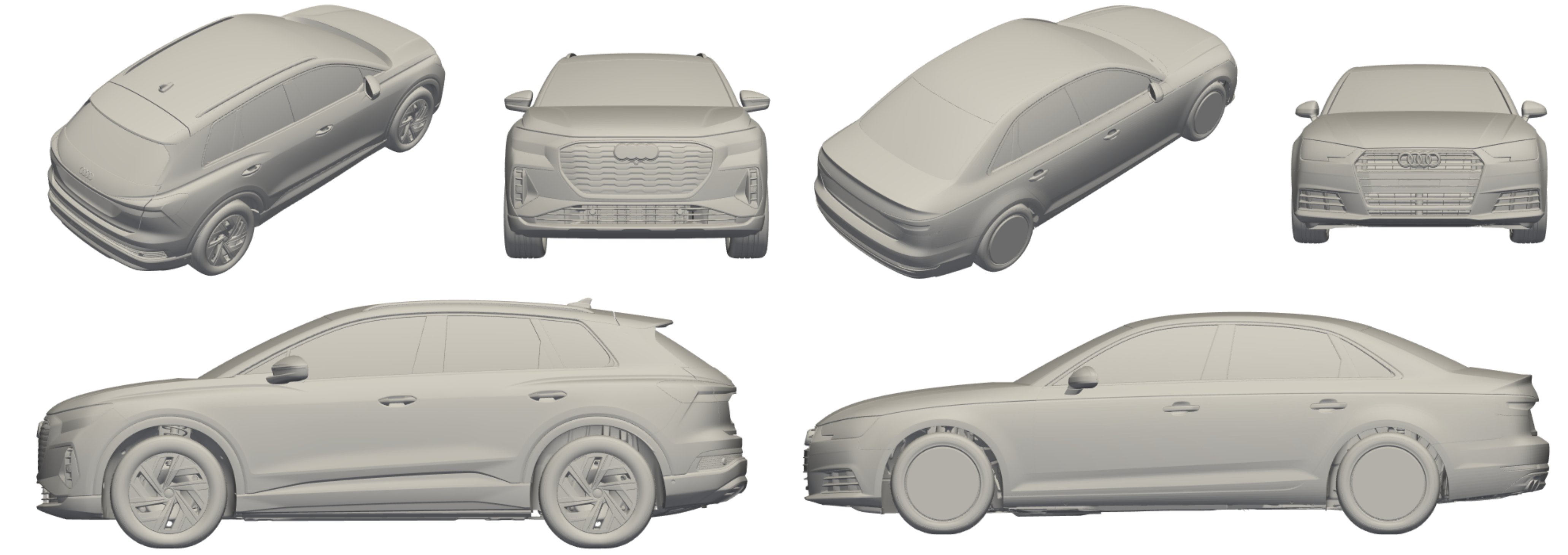}
    \caption{Two example geometries similar to the dataset's geometries. The dataset consists of cars from different projects and car types. We cannot show the exact geometries from the training set for confidentiality reasons.}
    \label{fig:example_geometries}
\end{figure}

We benchmark the methods using a real-world dataset, a collection of simulations aerodynamicists performed during ongoing iterations with stylists, resulting in geometries with diverse free-form changes from both aerodynamicists and stylists. Figure~\ref{fig:example_geometries} shows two geometries similar to those used in the dataset.

In contrast to datasets in previous studies \cite{Jacob.2022, GNN.Elrefaie.3122024, Data.Elrefaie.6142024, drivaerAshton2024} that generate geometries by parameterizing a limited set of similar baseline geometries, our dataset includes 32 baseline geometries from five different car projects, capturing the diversity present in the vehicle design process. Here, we designate specific training samples as baseline geometries: the starting points from which multiple variations are generated. A baseline geometry and its associated variations form a 'baseline group.' Typically, the overall geometric changes are largest between projects, moderate between baseline groups in a project, and smallest within a single baseline group. Hereafter, changes between baseline groups are referred to as inter-baseline group changes, while changes within a single baseline group are referred to as intra-baseline group changes.

\begin{figure}[!htbp]
    \centering
    \includegraphics[width=\linewidth]{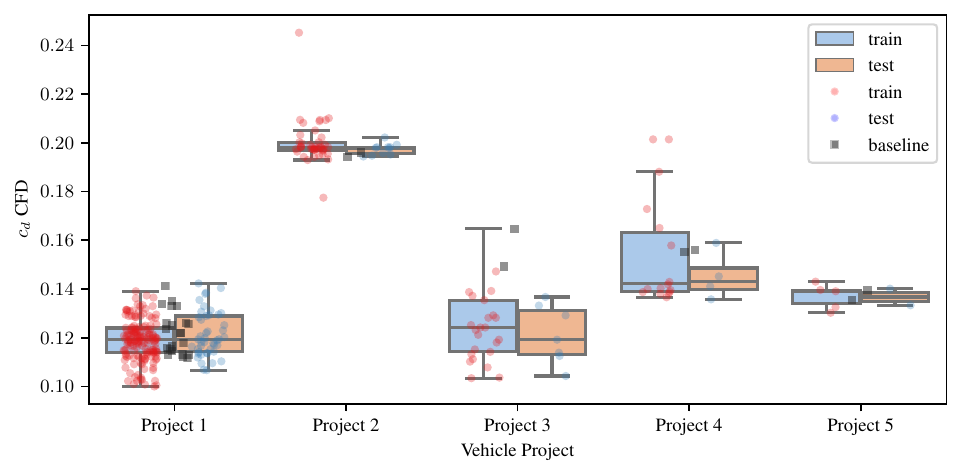}
    \caption{Distribution of $c_d$ across all projects. The samples are colored according to the split and whether they represent a baseline geometry. A baseline geometry is the initial reference geometry from the training split, used as a starting point for iterative modifications to evaluate design variations. Note: We scaled the $c_d$ values by subtracting a constant such that the sample with the lowest $c_d$ has a value of 0.1. This scaling is applied consistently across all plots displaying absolute $c_d$.}
    \label{fig:data distribution}
\end{figure}

Our dataset consists of 343 samples, 274 of which were used for training and 69 exclusively for testing. The dataset contains a mix of simulations with and without underhood flow. All simulations were based on Detached-Eddy Simulations performed using OpenFOAM. Islam et al. \cite{Islam.2009} validated the simulation methodology against wind tunnel measurements, and the setup is described in the dissertation by Sebald \cite{SimulationSetup.dissertation}. Each simulation takes about 20 hours using 960 CPU cores; we simulate for 4 seconds physical time and average the  $c_d$ from the last two seconds. Table~\ref{tab:dataset} shows the number of samples in each project, and Figure~\ref{fig:data distribution} illustrates the distribution of the $c_d$ across different projects and splits. 

\begin{table}[ht]
    \centering
    \begin{tabular}{p{2cm} p{2.8cm} p{2.8cm} p{4cm}}
        \hline
        \makecell[tl]{\textbf{Vehicle} \\ \textbf{Project}} & 
        \makecell[tl]{\textbf{Number of} \\ \textbf{Train Samples}} & 
        \makecell[tl]{\textbf{Number of} \\ \textbf{Test Samples}} & 
        \makecell[tl]{\textbf{Underhood Flow}} \\
        \hline
        Project 1 & 179 & 45 & No \\
        Project 2 & 45  & 11 & Yes \\
        Project 3 & 25  & 7  & No \\
        Project 4 & 18  & 4  & Train: Yes (5) / No (13); Test: No \\
        Project 5 & 7   & 2  & Yes \\
        \hline
    \end{tabular}
    \caption{A summary of the number of samples across projects for train and test splits and an indication of whether the simulations were performed with underhood flow.}
    \label{tab:dataset}
\end{table}

\section{Methodology}
\label{sec:Methodology}

\subsection{Preprocessing}
\label{section:preprocessing}

We choose a bounding box slightly larger than the largest car in our dataset. Each car is then centered within this bounding box. We then generate the SDF on a Cartesian grid with a size of 128 x 32 x 32. The value of the SDF (Equation \ref{eq:sdf}) in each grid cell is the distance between the cell center and the closest point on the surface of the geometry; the sign is positive for the points within the geometry and negative if outside. We choose a resolution of 128 x 32 x 32 to balance the computational cost and accuracy. We compute the SDF using OpenVDB \cite{OpenVDB}, which is faster and less noisy compared to our earlier approach \cite{Jacob.2022}. 

\begin{gather}
    \text{SDF}(x) = \operatorname{sign}(x) \min_{\forall p \in Z} |p - x|
    \label{eq:sdf}
\end{gather}

\noindent
where:

\noindent \( x \in \mathbb{R}^3 \) = a cell center of the Cartesian grid

\noindent \( Z \) = set of all points \( z \in \mathbb{R}^3 \) on the surface of the geometry

\noindent \( \operatorname{sign}(x) = -1 \) if \( x \) is inside the geometry, \( +1 \) otherwise

For the GNN-based method, we re-mesh \cite{remeshing.Valette.2008} the geometries to a coarser mesh of about 50,000 nodes, to keep the computational cost reasonable. For the GNN-based method, a finer mesh would have been beneficial, but we were limited to 50,000 nodes due to the available GPU memory.

\subsection{Data augmentation}

\begin{figure}[!htb]
    \centering
    \includegraphics[width=\linewidth]{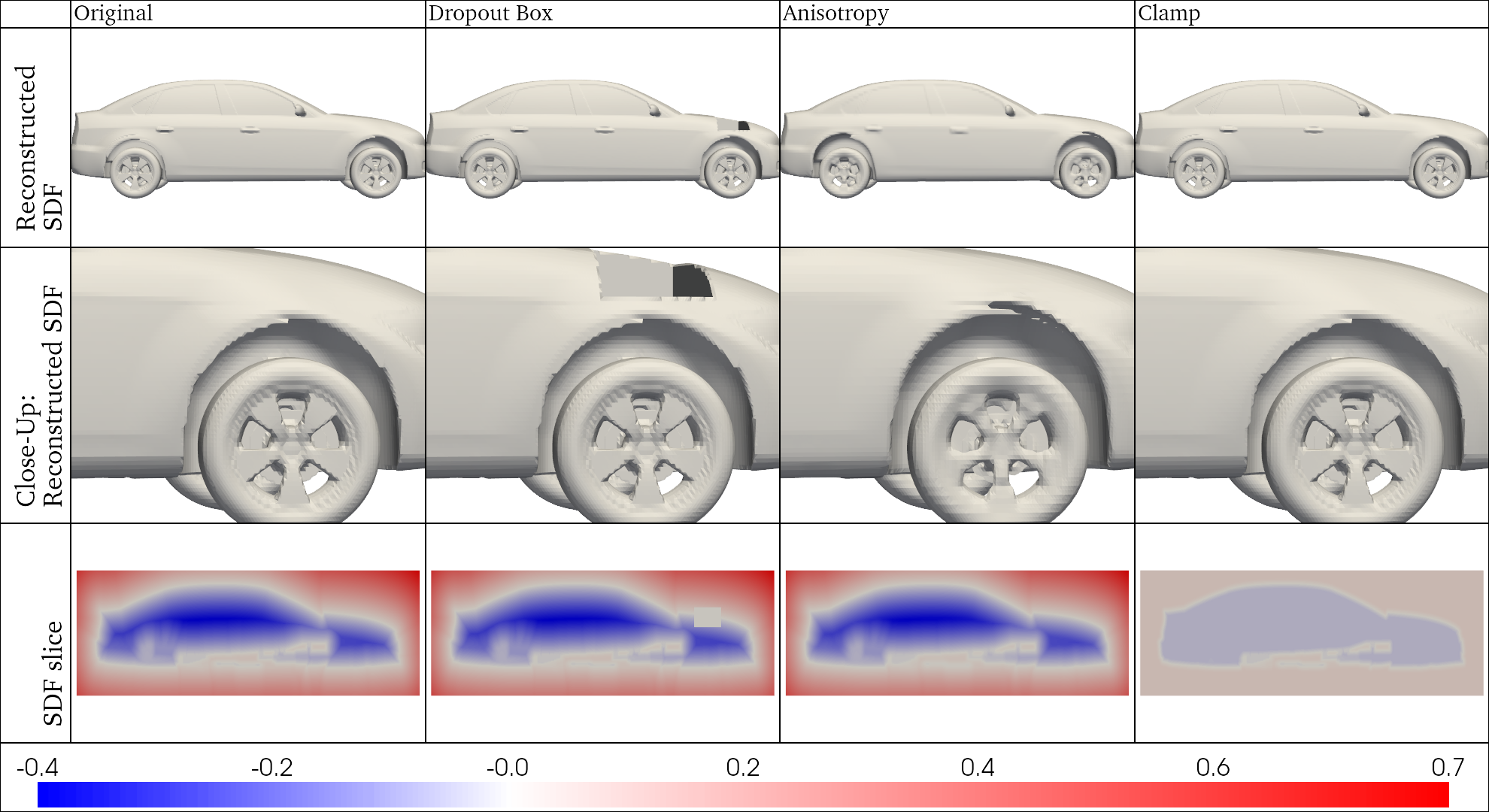}
    \caption{Examples of augmentation performed on a DrivAer geometry from the dataset \cite{drivaerAshton2024}. The rows, in order, show (1) reconstructed geometry from a Signed Distance Field (SDF), (2) a close-up of the reconstructed geometry, and (3) a slice of the SDF with the color bar below it. The columns, in order, show (1) the original geometry, (2) dropout box augmentation (randomly dropping out 3D regions), (3) anisotropy resampling augmentation (downsampling and upsampling the SDF), and (4) the clamp augmentation (clipping SDF values). We chose extreme values for some augmentation methods for demonstration purposes, as most augmented samples used for training exhibit little to no noticeable visual change.}
    \label{fig:augmentation_plot}
\end{figure}

Given the limited training data, we employ online data augmentation during training for the CNN-based method. In contrast to computationally expensive offline augmentation \cite{Jacob.2022}, which requires meshing geometries with multiple meshing methods and parameters before SDF generation, our online augmentation is applied dynamically during training, significantly reducing the preprocessing and storage costs. We consider the following objectives when developing the data augmentation method: 1) preventing the model from over-relying on specific regions, as suggested by the occlusion sensitivity performed in our earlier work \cite{Jacob.2022}; 2) ensuring invariance to minor geometric shifts; 3) improving robustness to noise from the SDF generation. Our online augmentation methods include:

\begin{enumerate}
    \item Clamp: We clip the SDF values to simulate narrow-band SDF using a threshold equal to the maximum SDF value multiplied by a random factor between 1e-5 and 1.
    \item Affine transformations: We slightly translate the entire SDF by up to 4\% along the X-axis to improve the translation invariance. 
    \item Noise injection: We add different types of random noises to the SDF, for example, Gaussian noise of varying strengths, to improve the robustness of noise from SDF generation.
    \item Elastic deformation: We apply small random deformations to the SDF. We set the parameters so that there is no significant visible deformation.
    \item Anisotropy resampling: We randomly downsample the SDF by a random factor between 1.2 and 2, then upsample it to its original resolution.
    \item Dropout box: We zero out multiple random 3D boxes with sizes between 1-5\% of the SDF volume to encourage learning redundant features.
\end{enumerate}

Figure~\ref{fig:augmentation_plot} illustrates examples of a few augmentation methods. The augmentations are performed concurrently on CPUs during training and have little to no impact on the training cost. We randomly apply augmentations to 75\% of the samples in a batch with randomly varying strengths. Based on experiments performed on the DrivAer dataset (in a setup similar to \cite{Jacob.2022}), we observed that the augmentations collectively improve the model's generalization, and our proposed online augmentation method performs on par with the offline augmentation method while significantly reducing the preprocessing cost. 

\subsection{Model architecture}

\begin{figure}[!htbp]
    \centering
    \includegraphics[width=\linewidth]{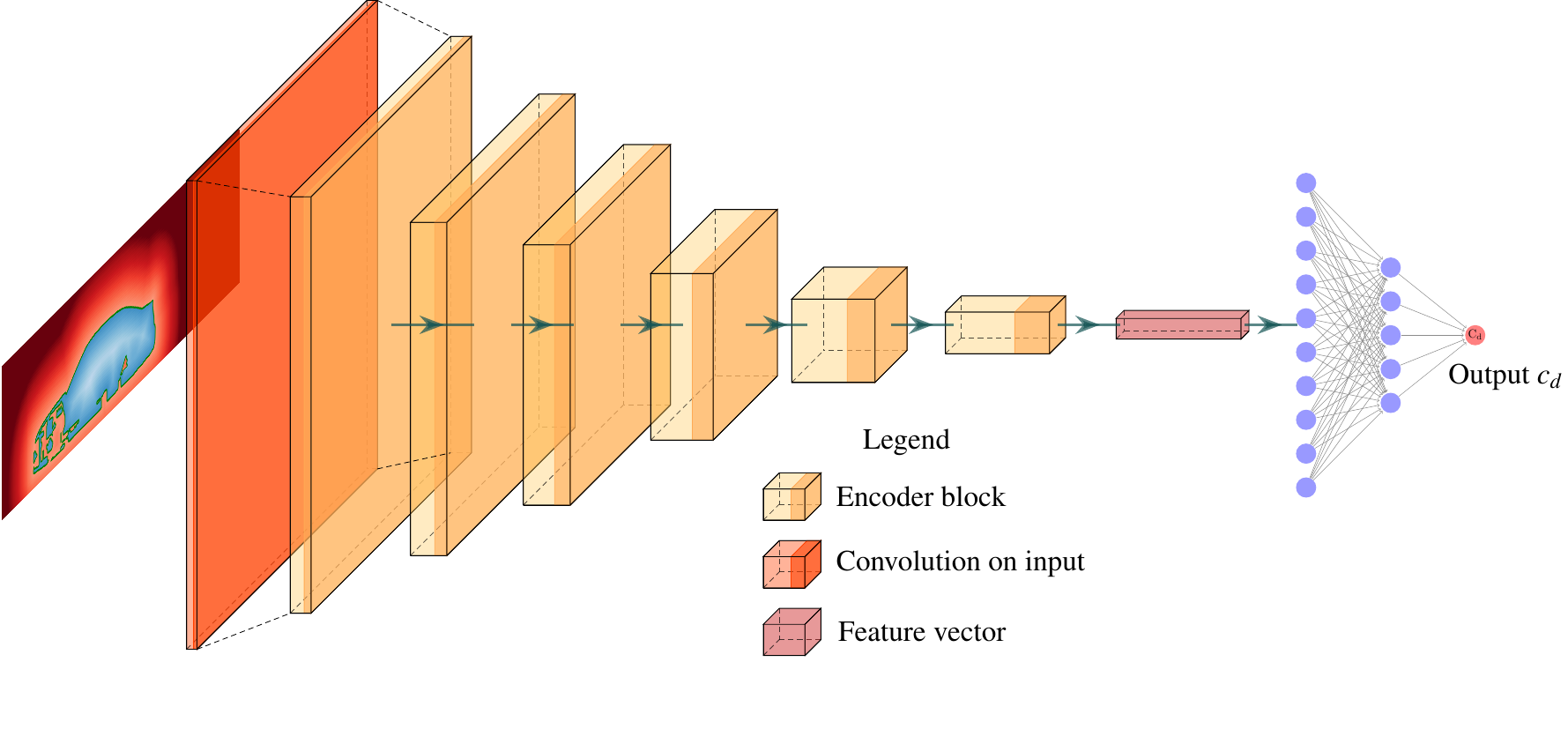}
    \caption{CNN-based method architecture illustration. We use a CNN-based encoder that takes a signed distance field as input and passes it through an input convolution layer, followed by a series of encoder blocks. The extracted feature vector is then passed through a series of fully connected layer blocks to predict $c_d$. We created the architecture illustration using PlotNeuralNet \cite{PlotNeuralNet} and adapted with permission from Jacob et al. \cite{Jacob.2022}. © 2022 SAE International.}
    \label{fig:UNet architecture}
\end{figure}

The CNN-based method takes the SDF as input and uses a CNN-based encoder. The feature vector from the encoder is then passed through a series of fully connected layer blocks to predict the $c_d$. The encoder has an input convolution layer followed by six encoder blocks. Each encoder block has, in order, a (concurrent spatial and channel) squeeze and excitation layer \cite{Squeeze_and_excitation.Hu.2018, Squeeze_and_excitation.Roy.2018}, activation, (dilated) convolution layer, group normalization \cite{groupnormalization.2018}, and a dropout layer. The extracted feature vector is first passed through a multi-head attention layer, followed by two fully connected layer blocks. Each fully connected layer block has an activation, batch normalization, and a fully connected layer. The squeeze and excitation layers and the multi-head attention layer help the model focus on the most relevant information. We apply standard scaling on the inputs and outputs, use RAdam optimizer \cite{radam.liu2019} with a cyclic learning rate scheduler \cite{CLR.2017}, and for additional normalization, we use stochastic depth \cite{stochasticdepth}. The CNN-based method has approximately 2 million trainable parameters. 

For the GNN-based method, we use a commercial tool that uses GNN, and the model was trained by experts from the firm that developed the commercial tool. The GNN-based method has approximately 5 million trainable parameters. 

\subsection{Evaluation metrics}
We use Mean Absolute Error (MAE – Equation~\eqref{eq:mae}) as the loss function and MAE and Maximum Absolute Error (MaxAE – Equation~\eqref{eq:maxae}) as the metrics for evaluating the test samples. While MAE provides an overall error measure, MaxAE represents the largest absolute error observed. For further evaluation, we evaluate the Direction Prediction Accuracy (DPA - Equation~\eqref{eq:direction_prediction_accuracy}) of the drag deltas (True drag delta, Equation~\eqref{eq:delta_cd}, Predicted drag delta, Equation~\eqref{eq:delta_cd_pred}).

\begin{equation}
    \text{MAE} = \frac{1}{n_{\text{test}}} \sum_{i=1}^{n_{\text{test}}} \left| c_d^i - \hat{c}_d^i \right|
    \label{eq:mae}
\end{equation}

\begin{equation}
    \text{MaxAE} = \max_{i \in \{1, \dots, n_{\text{test}}\}} \left( \left| c_d^i - \hat{c}_d^i \right| \right)
    \label{eq:maxae}
\end{equation}

\begin{equation}
    \Delta c_d^i = c_d^i - \bar{c}_d^j
    \label{eq:delta_cd}
\end{equation}

\begin{equation}
    \Delta \hat{c}_d^i = \hat{c}_d^i - \bar{c}_d^j
    \label{eq:delta_cd_pred}
\end{equation}

\begin{equation}
    \text{DPA} = 100 \times \frac{1}{n_{\text{test}}} \sum_{i=1}^{n_{\text{test}}} \left[ \text{sign}\left(\Delta c_d^i\right) = \text{sign}\left(\Delta \hat{c}_d^i\right) \right] \quad [\%]
    \label{eq:direction_prediction_accuracy}
\end{equation}

\noindent
where:

\noindent \( n_{\text{test}} \) = Number of test samples

\noindent \( c_d^i \) = True drag coefficient

\noindent \( \hat{c}_d^i \) = Predicted drag coefficient

\noindent \( \bar{c}_d^j \) = True drag coefficient of the corresponding baseline geometry

\noindent \( \Delta c_d^i \) = True change in drag coefficient relative to the baseline geometry

\noindent \( \Delta \hat{c}_d^i \) = Predicted change in drag coefficient relative to the baseline geometry

\noindent \(\left[ \cdot \right]\) = evaluates to 1 if the contained condition is true and 0 otherwise

\noindent \(\text{sign}(\cdot)\) = sign function.

\section{Results}
\label{sec:Results}

\begin{figure}[!htbp]
    \centering
    \includegraphics[width=\linewidth]{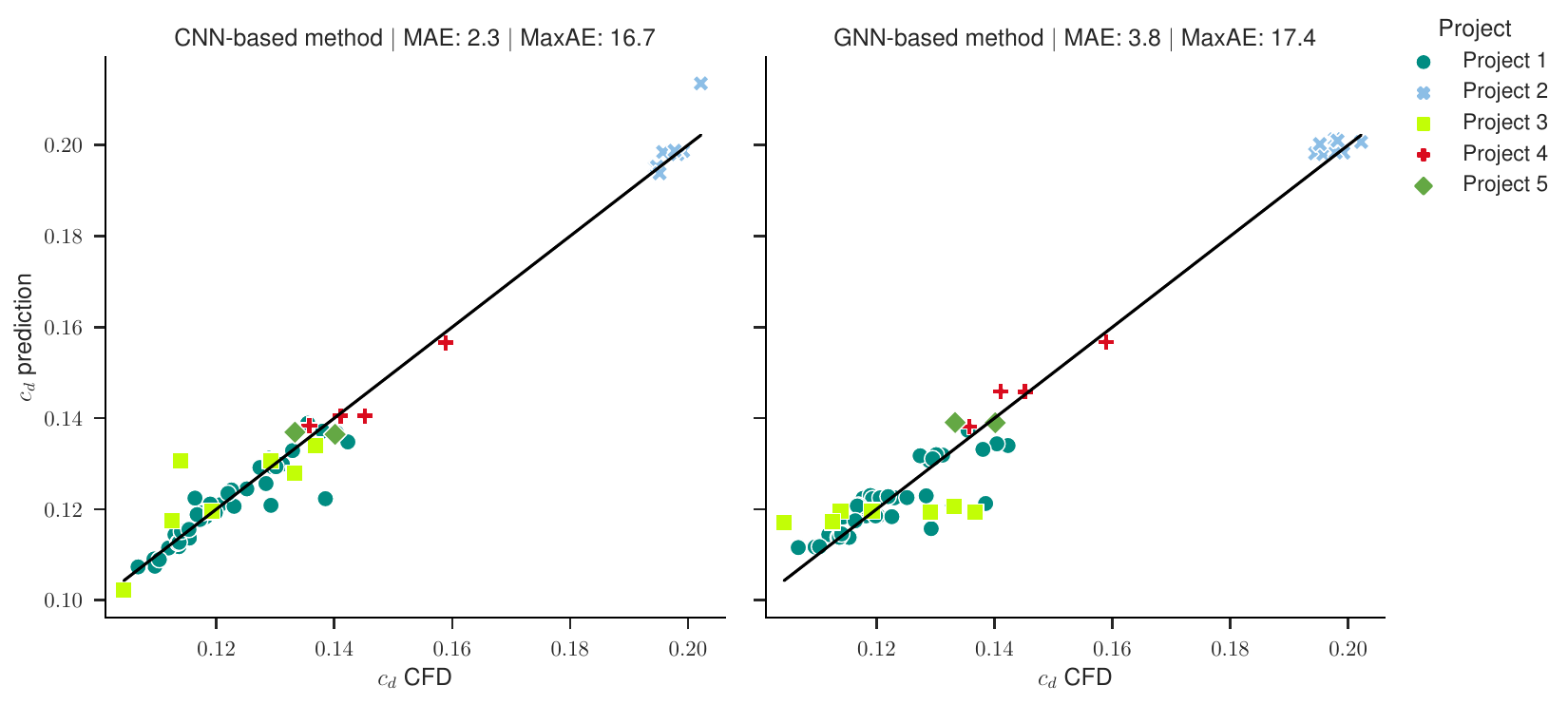}
    \caption{Correlation plots between the true $c_d$ and predicted $c_d$ for test samples for both methods (left: CNN-based method; right: GNN-based method). The subtitle has the overall Mean Absolute Error (MAE) and Maximum Absolute Error (MaxAE) for that method in drag counts, and one drag count = 0.001 drag.}
    \label{fig:correlation plot}
\end{figure}

For both methods, we train a single model for all the vehicle projects (Table \ref{tab:dataset}). We take this approach because it is common in real-world scenarios where the number of samples in some projects is relatively small (for example, projects 4 and 5 in this dataset). When training with all available projects, the prediction accuracy in the smaller projects is often better due to similarities with the other projects and helps avoid overfitting. We have seen similar behavior on DrivAer datasets with different parameterizations in our earlier work \cite{Jacob.2022}. Alternatives to this approach include training a model for individual projects and training one model per car type.

\begin{table}[H]
    \centering
    \begin{tabular}{p{1.5cm} p{1.5cm} p{1.5cm} p{1.5cm} p{1.5cm} p{1.5cm} p{1.5cm}}
        \hline
        \makecell[tl]{\textbf{Project} \\ \textbf{number}} & 
        \makecell[tl]{\textbf{CNN} \\ \textbf{MAE} \\ \textbf{(drag} \\ \textbf{count)}} & 
        \makecell[tl]{\textbf{GNN:} \\ \textbf{MAE} \\ \textbf{(drag} \\ \textbf{count)}} & 
        \makecell[tl]{\textbf{CNN:} \\ \textbf{MaxAE} \\ \textbf{(drag} \\ \textbf{count)}} & 
        \makecell[tl]{\textbf{GNN:} \\ \textbf{MaxAE} \\ \textbf{(drag} \\ \textbf{count)}} & 
        \makecell[tl]{\textbf{CNN:} \\ \textbf{DPA} \\ \textbf{(\%)}} & 
        \makecell[tl]{\textbf{GNN:} \\ \textbf{DPA} \\ \textbf{(\%)}} \\
        \hline
        1 & 1.9 & 3.3 & 16.2 & 17.2 & 76 & 76 \\
        2 & 1.8 & 2.6 & 11.4 & 5.0 & 82 & 73 \\
        3 & 4.8 & 9.1 & 16.7 & 17.4 & 100 & 100 \\
        4 & 2.5 & 2.5 & 2.5 & 4.8 & 100 & 100 \\
        5 & 3.6 & 3.4 & 3.6 & 5.7 & 0 & 0 \\
        Overall   & 2.3 & 3.8 & 16.7 & 17.4 & 78 & 77 \\
        \hline
    \end{tabular}
    \caption{Test metrics for CNN-based and GNN-based methods. We compare the overall and project-wise Mean Absolute Error (MAE), Maximum Absolute Error (MaxAE), and directional prediction accuracy (DPA). The directional prediction accuracy is the percentage of test samples where the model correctly predicts whether the $c_d$ increases or decreases relative to the baseline geometry. MAE and MaxAE are represented in drag counts, and one drag count = 0.001 drag.}
    \label{tab:metrics}
\end{table}

The correlation plots (Figure~\ref{fig:correlation plot}) indicate that both methods can make reasonable predictions except for some samples or projects. Notable differences between the methods, as observed from the correlation plot, and metrics table \ref{tab:metrics} include: (1) In project one, the GNN-based method has higher errors for a few more samples than the CNN-based method, (2)  For project two the overall error for the GNN-based method is higher than the CNN-based method, but the CNN-based method has a much higher error for one sample, (3) In project three the CNN-based method is able to capture the overall correlation except for one sample, but the GNN-based method almost predicts a constant for all the samples. 

\subsection{Analyzing trends}

\begin{figure}[!htb]
    \centering
    \includegraphics[width=\linewidth]{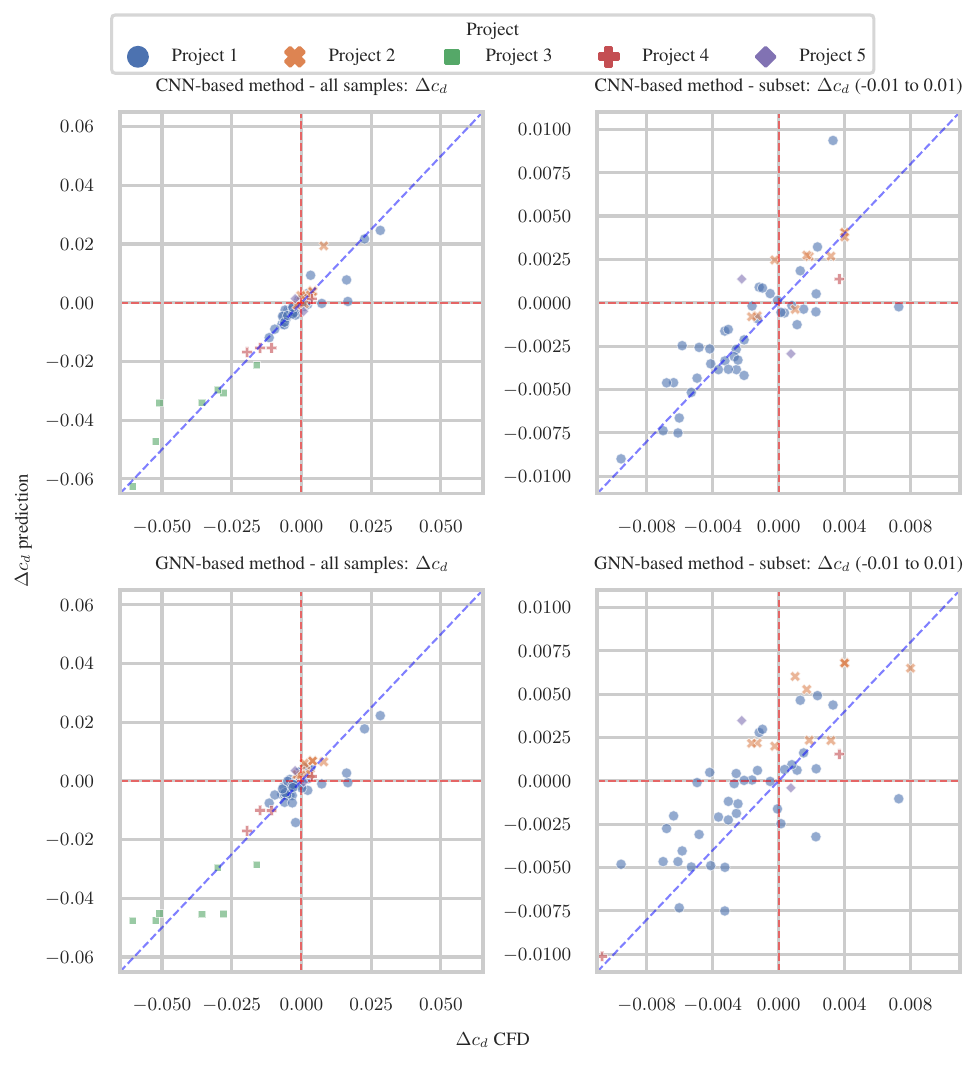}
    \caption{Delta plot: visualizing directional prediction accuracy (DPA). One objective for aerodynamicists is determining whether a geometric change increases or reduces $c_d$ compared to the baseline geometry. This plot compares the models' ability to predict the correct direction of $\Delta c_d$ (i.e., change in $c_d$ relative to the baseline geometry). The X-axis shows $\Delta c_d$ from CFD, and Y-Axis shows $\Delta c_d$ from model predictions. The first row has CNN-based method predictions, and the second row has the GNN-based method predictions. The first column shows all test samples, while the second column zooms into a subset region. Test samples falling in quadrants one or three indicate a correctly predicted direction, representing agreement in the direction of $\Delta c_d$ between simulation and model predictions. Both methods can predict the correct $\Delta c_d$ for a similar percentage of samples (see Table \ref{tab:metrics} for a detailed breakdown), and prediction error occurs primarily when the $\Delta c_d$ values are small.}
    \label{fig:delta plot}
\end{figure}

\begin{figure}[!htb]
    \centering
    \includegraphics[width=\linewidth]{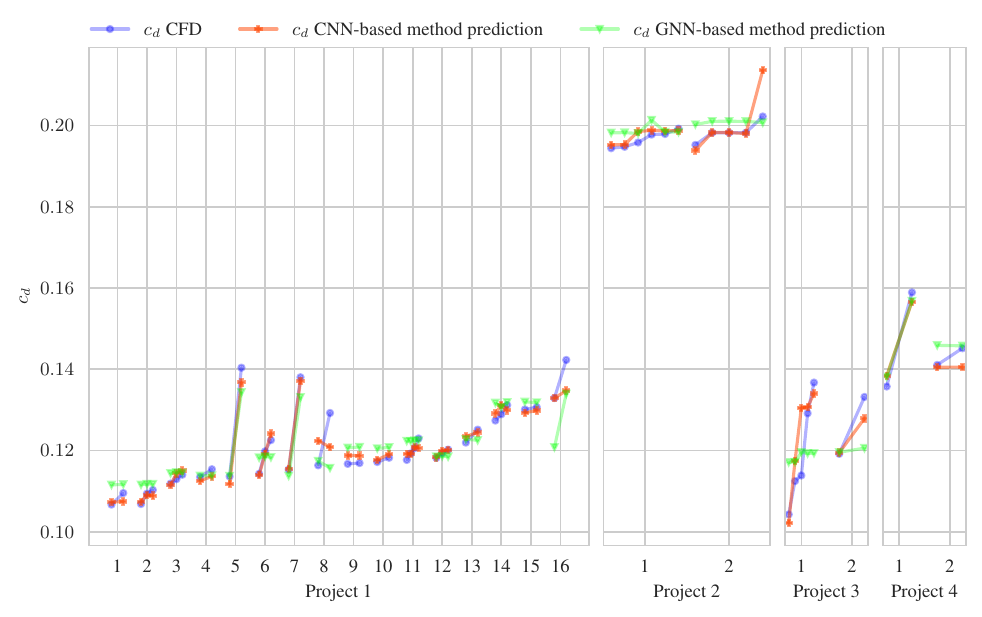}
    \caption{The plot compares trends in $c_d$ values from CFD and model predictions across baseline groups for test samples. Each baseline group consists of variations of a single baseline geometry; only baseline groups with at least two test samples are included. The X-axis represents projects, with numbers indicating individual baseline groups within each project, and the Y-axis represents $c_d$. Both methods capture the overall inter-baseline group trends but do not effectively capture intra-baseline group trends. This shortcoming is particularly notable in the GNN-based method, which frequently predicts nearly constant $c_d$ for many baseline groups.}
    \label{fig:trends plot}
\end{figure}

For the aerodynamicists, it is vital to predict the right direction of the $c_d$ change compared to the baseline geometry and to predict $c_d$ trends similar to the simulation for intra-baseline group changes. To evaluate these criteria, we use a delta plot (Figure~\ref{fig:delta plot}) to compare the direction of $c_d$ prediction and a trends plot (Figure~\ref{fig:trends plot}) to evaluate the $c_d$ prediction trends.

From the delta plot (Figure~\ref{fig:delta plot}), we can see that most of the samples for which the methods make a mistake in the direction have a $\Delta c_d$ lower than 0.007. These errors could be due to minor geometric feature changes compared to the baseline, which are subtle enough that the model struggles to account for them. The metrics table (Table~\ref{tab:metrics}) shows that both methods successfully predict the correct direction for most samples. Over all test samples, the CNN-based method achieves 78\% DPA, while the GNN-based method achieves 77\% DPA.

Analyzing $c_d$ prediction trends across baseline groups (Figure~\ref{fig:trends plot}) reveals distinct prediction differences between the methods. In cases where baseline groups exhibit huge $c_d$ variations — in some instances (e.g., project 1: groups 5 and 7), both the models capture the overall trend effectively, yet in other instances (e.g., project 1: groups 8 and 16), both the models fall short. For the baseline groups with many test samples (e.g., both groups in project 2), the CNN-based method tracks the trends more accurately than the GNN-based method, which mostly predicts constants within a baseline group. Most baseline groups have either few test samples or relatively small $c_d$ changes; here, the GNN-based method mostly defaults to a constant output (e.g., project 1: groups 1 and 9), while the CNN-based method, although similar in many cases, is relatively more accurate (e.g., project 1: groups 2, 10 and 12) at capturing the subtle intra-baseline group trends. 

Overall, while both methods capture the inter-baseline group trends well, both methods struggle to different extents to account for the finer intra-baseline group trends. Relatively, the geometric changes within the intra-baseline groups are smaller compared to the inter-baseline groups, which suggests that both methods find it challenging to capture smaller variations in the geometry; this could be due to the insufficient resolution of the SDF for the CNN-based method and the mesh size for the GNN-based method. On the other hand, many baseline groups only have a few training samples, which could lead to a few exotic or unique geometries. Given the dataset's modest size for deep learning applications, expanding the training data could also improve the prediction accuracy.

\subsection{Performance}

We evaluate the performance of both models on a machine with a single NVIDIA Tesla V100 GPU. Table~\ref{tab:benchmark_values} summarizes both models' training, preprocessing, and prediction times. Although both methods achieve prediction times under one second, the time taken for preprocessing (i.e., SDF generation or coarser surface mesh generation) a new geometry takes about a minute, excluding the time taken to load the geometry. The time taken to load a geometry is variable and depends on various factors, including mesh size, storage format, and the speed of the storage medium. For example, it takes about 40 seconds to load a mesh with approximately 10 million nodes from an SSD. Further optimization to improve overall prediction time is possible, for example, by performing preprocessing using GPU. Currently, for a new geometry, predictions can be made in under two minutes when utilizing fast storage, enabling near real-time predictions. Compared to the simulation wall clock time of about 20 hours (using 960 CPU cores), the model's total prediction time on a CPU is at least 600 times faster. Furthermore, relatively low training times allow frequent retraining when new data becomes available.

\begin{table}[ht]
    \centering
    \begin{tabular}{p{3.5cm} p{2.2cm} p{3.0cm} p{2.5cm}}
        \hline
        \textbf{Method} 
        & \textbf{Training Time [h]} 
        & \textbf{Preprocessing Time [s]} 
        & \textbf{Prediction Time [s]} \\
        \hline
        GNN-based method & $\sim$9.5 & $\sim$60 & $<$1 \\
        CNN-based method & $\sim$3 & $\sim$30 & $\sim$0.1 \\
        \hline
    \end{tabular}
    \caption{Comparison of GNN-based and CNN-based methods in terms of training, preprocessing, and prediction times. We use one NVIDIA Tesla V100 to train both models while preprocessing and prediction are performed on a CPU. Note: Preprocessing time excludes data loading into memory.}
    \label{tab:benchmark_values}
\end{table}

In real-world applications, it is essential to assess the cost of simulations when considering the creation of surrogate models. Generating training data is the most computationally expensive part of training surrogate models; the simulations in this study took over 5 million CPU hours. However, all simulations were created during regular vehicle development projects.

\section{Conclusion}
\label{sec:Conclusion}
In this study, we use a real-world automotive dataset to benchmark two surrogate modeling approaches for aerodynamic $c_d$ prediction: a Convolutional Neural Network (CNN)- based method that uses Signed Distance Fields (SDF) as input and a Graph Neural Network (GNN)- based commercial tool that directly works with a mesh. Our dataset was curated from high-fidelity simulations during routine vehicle development, which bridges the gap between academic datasets and industrial use cases, and we demonstrate that data-driven surrogate models can be used effectively to predict the $c_d$ during vehicle development. 

Both methods achieved reasonable accuracy in predicting the $c_d$ and, more importantly, predicting the correct direction of $c_d$ change relative to the baseline geometry. The CNN-based method, with its lower training cost, achieved an MAE of 2.3 drag counts and a directional prediction accuracy of 78\%, whereas the GNN-based method had a higher MAE of 3.8 drag counts and a directional prediction accuracy of 77\%. Both methods can predict the $c_d$ of a new geometry in under two minutes by using only the CPU, which is at least 600 times faster than simulations. The results indicate that both methods can help aerodynamicists find promising geometric variations before committing to time-consuming simulations. The choice of method should depend on the use case and the advantages and disadvantages of the methods. For example, in cases with limited computing resources, the CNN-based method is better suited, and in cases with noisy/unclean geometries or when surface field predictions are desired, GNN-based methods fit better.

Analyzing the trends of baseline groups revealed that both methods are adept at capturing the broader inter-baseline group trends but struggle to capture the subtler intra-baseline group trends. We hypothesize these challenges are due to the loss of information during the preprocessing stages (i.e., discretization for the CNN-based method and mesh-down sampling for the GNN-based method) and the relatively small size of the training dataset. Furthermore, in this study, the available GPU memory was a limitation of the mesh size for the GNN-based method. 

As aerodynamicists often make small geometric changes that can significantly affect $c_d$, improving the models to capture small geometric variations is essential. Increasing the input resolution could reduce the information lost during preprocessing; there have been some recent studies in this direction. For example, sparse convolution \cite{fVDB} enables training with higher resolution SDFs at reduced compute costs. Additionally, recent studies \cite{neuraloperator.Domino, GNN.Nabian.11262024} have explored training GNNs with larger meshes. Another potential direction is the development of hybrid multi-model methods that use both SDF and mesh as input; this can lead to models that leverage the respective strengths of the methods; the SDF branch focuses on capturing global information and predicting the volumetric fields, while the GNN focuses on the surface-level details and predicting the surface fields. 

\section{Acknowledgements}
The authors acknowledge the assistance of Dr. Jutta Pieringer and colleagues, who collected the dataset and helped us throughout the study. 

\section*{Declaration of generative AI and AI-assisted technologies in the writing process}
During the preparation of this work the authors used ChatGPT in order to improve the readability of the text. After using this tool, the authors reviewed and edited the content as needed and take full responsibility for the content of the publication.

\bibliographystyle{elsarticle-num} 
\bibliography{references}

\section*{Disclaimer}
The results, opinions and conclusions expressed in this publication are not necessarily those of Volkswagen Aktiengesellschaft. 

\end{document}